\documentclass{article}
\usepackage{ijcai20}

\usepackage{times}
\usepackage{soul}
\usepackage{url}
\usepackage[hidelinks]{hyperref}
\usepackage[utf8]{inputenc}
\usepackage[small]{caption}
\usepackage{graphicx}
\usepackage{amsmath, bm}
\usepackage{amsthm}
\usepackage{color, soul}
\usepackage{amsfonts}
\usepackage{multirow}
\usepackage{booktabs}
\usepackage{algorithm}
\usepackage{algorithmic}
\setul{0.5ex}{0.2ex}
\setulcolor{blue}

\title{Keyphrase Prediction With Pre-trained Language Model}

\author{
Rui Liu$^{1,2}$
\and
Zheng Lin$^1$\and
Weiping Wang$^1$
\affiliations
$^1$Institute of Information Engineering, Chinese Academy of Sciences\\
$^2$School of Cyber Security, University of Chinese Academy of Sciences
\emails
\{liurui1995, linzheng, wangweiping\}@iie.ac.cn,
}

\begin{document}

\maketitle

\begin{abstract}

Recently, generative methods have been widely used in keyphrase prediction, thanks to their capability to produce both present keyphrases that appear in the source text and absent keyphrases that do not match any source text.
However, the absent keyphrases are generated at the cost of the performance on present keyphrase prediction, since previous works mainly use generative models that rely on the copying mechanism and select words step by step. Besides, the extractive model that directly extracts a text span is more suitable for predicting the present keyphrase.
Considering the different characteristics of extractive and generative methods, we propose to divide the keyphrase prediction into two subtasks, i.e., present keyphrase extraction (PKE) and absent keyphrase generation (AKG), to fully exploit their respective advantages.
On this basis, a joint inference framework is proposed to make the most of BERT in two subtasks.
For PKE, we tackle this task as a sequence labeling problem with the pre-trained language model BERT.
For AKG, we introduce a Transformer-based architecture, which fully integrates the present keyphrase knowledge learned from PKE by the fine-tuned BERT.
The experimental results show that our approach can achieve state-of-the-art results on both tasks on benchmark datasets.
\end{abstract}

\section{Introduction}

Keyphrase prediction aims to automatically obtain many condensed phrases or words, which can highly summarize the primary information of a document. A solution to this task is essential for numerous downstream NLP tasks, e.g., recommendation, information retrieval~\cite{DBLP:conf/sigir/UshikuMKT17}, and summarization~\cite{pasunuru-bansal-2018-multi}. In practical applications, people can quickly gain the required content from the Internet through keyphrases.

\begin{figure}[t]
\centering
\small
\begin{tabular} {|p{7.3cm}|}
\hline
\textbf{Document:}(total 173 words) \\
On the syntactic and functional correspondence between hybrid (or layered) normalisers and \textbf{\textsl{abstract machines}}. 
We show how to connect the syntactic and the functional correspondence for normalisers and \textbf{\textsl{abstract machines}} implementing hybrid (or layered) \textbf{\textsl{reduction strategies}}, ...
\ul{Many fundamental strategies in the literature are hybrid, in particular, many full reducing strategies ...}
If we follow the standard \textbf{\textsl{program transformation}} steps the ...
\ul{However, a solution is possible based on establishing the shape invariant of well formed continuation stacks.}
\ul{We illustrate the problem and the solution with the derivation of substitution based ...}
\ul{The machine we obtain is a substitution based, eval apply, open terms version of Pierre cregut's ...} \\
\specialrule{0.0em}{0pt}{1pt} 
\hline
\\
\specialrule{0.0em}{0pt}{-8pt}
\textbf{Present Keyphrases:} abstract machines; reduction strategies; program transformation \\
\textbf{Absent Keyphrases:} operational semantics; full reduction \\
\hline
\end{tabular}
\caption{A sample document with labeled keyphrases. The present keyphrases are shown in bold. There are no present keyphrases in the sentences underlined in blue.}
\label{fig1}
\vspace{-5mm}
\end{figure}

Existing keyphrase prediction approaches mostly focus on either extractive or generative methods. Extractive methods aim to select \textit{present keyphrases} (e.g., “abstract machines” in Figure~\ref{fig1}), which appear in the document. However, an issue of these methods is that they cannot produce \textit{absent keyphrases} (e.g., “operational semantics” in Figure~\ref{fig1}), which do not exist in the document.
Another line of works~\cite{DBLP:conf/acl/MengZHHBC17,DBLP:conf/aaai/ChenGZKL19} treats the keyphrase prediction as a sequence-to-sequence learning problem and uses the encoder-decoder framework to generate present and absent keyphrases simultaneously.
These works reveal that adopting the copy mechanism is more effective than directly generating words from the vocabulary. However, the copying mechanism generates a word at each time step and does not take dependencies between the selected words into consideration. 
Meanwhile, \citeauthor{chen-etal-2019-integrated}~\shortcite{chen-etal-2019-integrated} focused on improving the performance of the generative model with the assistance of an extractive model. Nevertheless, instead of directly extracting keyphrases from the original document, their proposed extraction model aimed to identify the importance of each word in the document, and the importance score was used to assist the generation of keyphrases. As a result, the potential of the extractive model has not been fully exploited.

To fully exploit the power of extraction and generation, we divide the keyphrase prediction problem into two processes: present keyphrase extraction (PKE) and absent keyphrase generation (AKG). 
For PKE, we address this work as a sequence labeling problem using a BiLSTM-CRF architecture; meanwhile, we employ the pre-trained model BERT~\cite{devlin-etal-2019-bert} to obtain the contextual embedding. 
Moreover, there are some sentences in the document that do not contain present keyphrases, and these noisy data can impair the performance of PKE. To tackle the issue above, we design a sentence filter module to select sentences that may contain present keyphrases to extract keyphrases more accurately.

For AKG, we exploit the extractive information from the shared BERT model fine-tuned on the PKG task.
Furthermore, the present keyphrase information provides an explicit summary of the topic of the article, and it can be used to guide the generation of absent keyphrases. 
To achieve this goal, we employ a Transformer-based model~\cite{DBLP:conf/nips/VaswaniSPUJGKP17} with the copying mechanism. 
Rather than simply using the find-tuned BERT as the encoder, we propose a gated fusion attention module, in which we use Transformer encoder representations to interact with the BERT representations. Afterwards, a gated fusion layer is introduced to fuse the present keyphrase knowledge and Transformer encoder representations.

The main contributions of this paper are listed as follows:
\begin{itemize}

\item We divide the keyphrase prediction task into two subtasks, i.e., PKE and AKG, to combine the advantages of both extractive models and generative models. To be specific, the extractive model thoroughly considers the dependencies between words to enhance the performance. Meanwhile, additional present keyphrase information enables the generation model to generate absent keyphrases that are close to the topic.

\item A shared BERT is utilized in the two subtasks (PKE and AKG) to benefit from the prior knowledge learned from data-rich corpora. Specifically, to take advantage of the BERT fine-tuned on PKE, we propose a gated fusion attention module to integrate the present keyphrase information in AKG. In this way, both the PKE and AKG tasks can be further refined through the pre-trained language model.

\item Experimental results on three benchmark datasets show that our model outperforms the state-of-the-art models significantly.
\end{itemize}

\section{Model}
\begin{figure*}[!t]
    \centerline{\includegraphics[width=0.65\textheight]{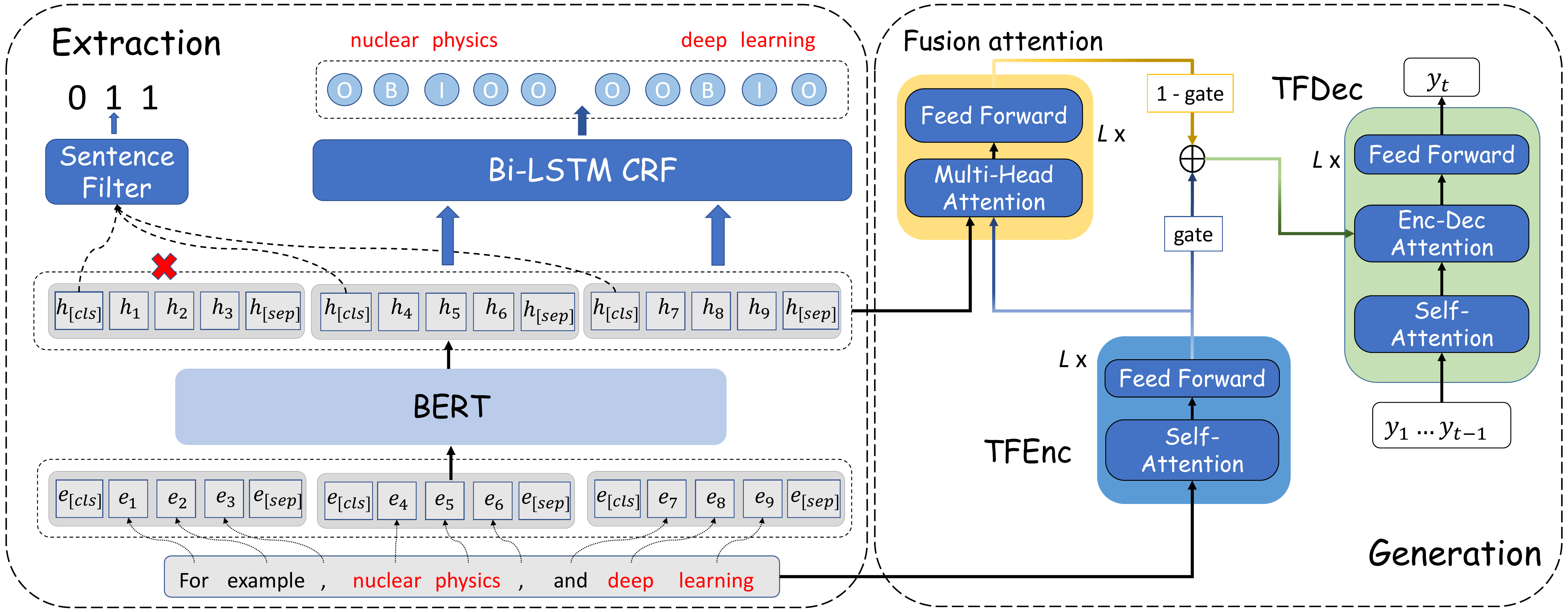}}
    \caption{The architecture of our proposed model. The $\bm{e}_i$ and $\bm{h}_i$ denote the embedding vector and the bert representation of the $i$-th word respectively. 
    Here, $y_t$ is the word predicted by the generative model at the $t$-th step according to the previously generated sequence.}
\label{fig2}
\end{figure*}

The structure of our model, including two submodels (i.e., BERT-PKE and BERT-AKG), is shown in Figure~\ref{fig2}. During training, we first train our labeling model into convergence. Then, the generative model is trained with BERT that is fine-tuned on our present keyphrase extraction task.
In testing, we jointly conduct two subtasks, a document is converted into hidden states via BERT encoder and Transformer encoder respectively, then we simultaneously extract present keyphrases and generate absent keyphrases.

\subsection{Problem Definition}
Given a document $\bm{x} = \{x_1, \ldots, x_{L_x}\}$, the goal is to obtain keyphrases $\mathcal{Y} = \{\bm{y}_1 \ldots, \bm{y}_M\}$ including present keyphrases and absent keyphrases, where $L_x$ is the length of the document, and $M$ is the number of keyphrases.

\subsection{Present Keyphrase Extraction}

The overall framework of our sequence labeling model (i.e., BERT-PKE) consists of two components: (i) BERT-based sentence filter, (ii) BiLSTM-CRF sequence labeling module. 

\subsubsection{BERT-based Sentence Filter}
Given a document $\bm{x}$, we first split it into some sentences $\mathcal{S}=\{sent_1, \ldots, sent_{L_s}\}$ by punctuation marks, where $L_s$ is the number of the sentences. To filter noisy sentences which do not contain present keyphrases, we add special tokens [CLS] and [SEP] at the start and the end of the sentence respectively, inspired by \cite{liu-lapata-2019-text}.
BERT encodes the processed document into contextual representations $\bm{H}$. The vector $\bm{h}_{[cls]}$ of the token [CLS] before each sentence is used as the sentence representation. We denote those representations as $\bm{G}^0=\{\bm{g}_i^0\}_{i=1}^{L_s}$, where $\bm{g}_i^0$ is the vector of the $i$-th sentence. 
The sentence representations $\bm{G}^0$ are then fed into 2-layer Transformer blocks. Each block contains two sub-layers: a multi-head self-attention layer and a fully connected feed-forward network:
\begin{equation}
  \bm{G}^l = \text{FFN}(\text{MultiHeadAtt}(\bm{G}^{l-1}, \bm{G}^{l-1}, \bm{G}^{l-1})).
\end{equation} 
The three inputs of the multi-head self-attention layer are query matrix, key matrix and value matrix from left to right. 
A residual connection is employed around each of the two sub-layers, followed by layer normalization. 

Afterwards, we obtain the confidence score for each sentence through a sigmoid function: 
\begin{equation}
    \bm{score} = \sigma(\bm{w}^T\bm{G}^2) \in \mathbb{R}^{L_s}.
\end{equation}
According to the sentence scores, we choose the top-\textit{K} candidate sentences for the subsequent sequence labeling process. Here, \textit{K} is set to 7, and we investigate the influence of the hyperparameter \textit{K} in section~\ref{impact}.
Each sentence is associated with a label $\widetilde{y}_i \in \{0, 1\}$, indicating whether the $sent_i$ contains any present keyphrase. We can train the sentence filter by minimizing the negative log-likelihood loss:
\begin{equation}
    \mathcal{L}_{f} = -\sum_{i=1}^{L_s}{\widetilde{y}_i\log{score_i}}.
\end{equation}
where $L_s$ is the number of the sentences in the document.

\subsubsection{Bi-LSTM CRF Sequence Labeling Architecture}

The contextualized vectors of the words from the selected sentences are fed into a BiLSTM to add sufficient expressive power. Then, a conditional random field (CRF) scores and labels the output of the BiLSTM network. 
As described in~\cite{DBLP:conf/naacl/LampleBSKD16}, given an input document $\bm{x}$, the score of the sequence of tag predictions $\bm{t}$ can be defined as:
\begin{equation}
    s(\bm{x}, \bm{t}) = \sum_{i=0}^n{\bm{A}_{t_i,t_{i+1}}} + \sum_{i=1}^n{\bm{P}_{i,t_i}},
\end{equation}
where $\bm{A}_{i,j}$ is the transition score from tag $i$ to tag $j$, and $\bm{P}_{i,j}$ is the score of the $j$-th tag of the $i$-th word.
The CRF model can be trained by minimizing the negative log-probability of the ground-truth tag sequence $\bm{t}$:
\begin{equation}
 \begin{split}
    \mathcal{L}_{c} &= -log(e^{s(\bm{x}, \bm{t})} / \sum_{\hat{\bm{t}} \in \bm{T}} e^{s(\bm{x}, \hat{\bm{t}})}) \\
    &= -s(\bm{x}, \bm{t}) + log(\sum_{\hat{\bm{t}} \in \bm{T}} e^{s(\bm{x}, \hat{\bm{t}})}).
 \end{split}
\end{equation}
The best sequence path can be found using the Viterbi decoding algorithm. 

In this work, rather than training the BiLSTM-CRF module on sentences selected by the sentence filter module, we train it exclusively on ground-truth positive sentences that contain present keyphrases.
This strategy removes most of the responsibility of content selection and allows the module to focus its efforts on labeling the document.
During testing, we first utilize the sentence filter to select top-\textit{K} sentences by calculating the confidence score. Thereafter, BiLSTM-CRF tags those selected sentences. Finally, we extract all the present keyphrases according to the IOB format~\cite{DBLP:conf/naacl/LampleBSKD16}.

The final loss of the overall extractive model BERT-PKE can be expressed as:
$\mathcal{L}_{PKE} = \mathcal{L}_{f} + \mathcal{L}_{c}$.

\subsection{Absent Keyphrase Generation}
The basic architecture of BERT-AKG is Transformer ~\cite{DBLP:conf/nips/VaswaniSPUJGKP17}, which consists of an encoder and a decoder.
Encoder and decoder both contain $L$-layer Transformer blocks. But the Transformer model has its vocabulary $\mathcal{A}$ and the words in this vocabulary are not tokenized by WordPiece. Besides, the BERT model we used is fine-tuned on the present keyphrase extraction task. To prevent the shared BERT from forgetting the knowledge of present keyphrases, we do not train the BERT model with the Transformer. In other words, we only treat the output vector of the shared BERT as a fixed supplementary knowledge to guide the generation procedure.

\subsubsection{Transformer Encoder with Fusion Attention Module}
Given a document $\bm{x}$, the Transformer encodes it into $\bm{U}$. We denote the word embedding of the document $\bm{x}$ as $\bm{U}^0=Embedding(\bm{x})$, and in the $l$-th layer:
\begin{equation}
  \bm{U}^l = \text{FFN}(\text{MultiHeadAtt}(\bm{U}^{l-1}, \bm{U}^{l-1}, \bm{U}^{l-1})).
\end{equation}
Meanwhile, the BERT model encodes $\bm{x}$ into representation $\bm{H}$. We adopt another multi-head attention module with \textit{L} layers to find the useful information of BERT representations which are conducive to generate absent keyphrases:
\begin{equation}
  \hat{\bm{U}}^l = \text{FFN}(\text{MultiHeadAtt}(\hat{\bm{U}}^{l-1}, \bm{H}, \bm{H})),
\end{equation} 
where $\hat{\bm{U}}^0 = \bm{U}^L$. 
Then, we use a soft gating weight to effectively merge the integrated BERT representation $\hat{\bm{U}}^L$ with the Transformer encoding representation $\bm{U}^L$. 
\begin{gather}
    gate = \sigma(\bm{W}_u[\bm{U}^L;\hat{\bm{U}}^L]), \\ 
    \bm{V} = gate \odot \bm{U}^L + (1-gate) \odot \hat{\bm{U}}^L,
\end{gather}
where $\bm{V}$ is the final encodings of the document $\bm{x}$, and $\odot$ is an element-wise multiplication.

\subsubsection{Transformer Decoder}
The Transformer decoder is also composed of a stack of $L$ identical layers. Except the self-attention sub-layer in the encoder module, each decoder layer contains another multi-head encoder-decoder attention sub-layer to perform attention over the output representation $\bm{V}$ of the encoder stack. We denote the input of the decoder as $\bm{D}^0$. Each decoder block is as follows:
\begin{gather}
     \bm{C}^l = \text{MultiHeadAtt}(\bm{D}^{l-1}, \bm{D}^{l-1}, \bm{D}^{l-1}), \\
    \bm{D}^l = \text{FFN}(\text{MultiHeadAtt}(\bm{C}^l, \bm{V}, \bm{V})),
\end{gather}
where, $\bm{D}^l$ is the output of the $l$-th decoder block.

To further improve the generation ability of the model, we incorporate the copying mechanism~\cite{see-etal-2017-get} with the Transformer decoder, where the attention distribution $\bm{a}_t$ from the last decoding layer indicates the probability of copying a word from the source text.  
Hence, the final predicted distribution $P$ at time step $t$ can be computed as:
\begin{equation}
P = p_{gen}P_{vocab} + (1-p_{gen})\sum_{i:w_i=w}a_t^i,
\label{eq:final}
\end{equation}
where $p_{gen} = \sigma(\bm{w}^T_d\bm{d}_t^L+b_t) \in [0,1]$ is a switch that controls the probability of generating a word from the vocabulary $\mathcal{A}$ or copying a word from the document, and $\bm{d}_t^L$ is $t$-step's output vector of the decoder.
The vocabulary distribution $P_{vocab}$ over the fixed vocabulary $\mathcal{A}$ is computed as $P_{vocab} = softmax(\bm{W}_v\bm{d}_t^L + \bm{b}_v).$ 

We train the generator by minimizing the cross entropy loss:
\begin{equation}
 \mathcal{L}_{AKG} = -\sum^{|\bm{y}|}_{t=1}{\log{P(y_t|\bm{y}_{1 \ldots t-1},\bm{x})}},
\end{equation}
where $y_t$ is the $t$-th word of keyphrase $\bm{y}$, and $|\bm{y}|$ is the length of ground-truth keyphrase $\bm{y}$.

\section{Experiment}

\subsection{Dataset}
We choose three datasets of scholarly documents for evaluation, which includes \textbf{KP20k}~\cite{DBLP:conf/acl/MengZHHBC17}, \textbf{NUS}~\cite{DBLP:conf/icadl/NguyenK07}, and \textbf{Krapivin}~\cite{krapivin2009large}. 
\textbf{KP20k} is a large-scale scholarly articles dataset with 528K articles for training, 20K articles for validation and 20K articles for testing. All the models are trained with the data from \textbf{KP20k}. We conduct zero-shot evaluations on the remaining two datasets following the previous work~\cite{DBLP:conf/acl/MengZHHBC17,DBLP:conf/aaai/ChenGZKL19}.
The statistics of the three datasets are shown in Table~\ref{tab4}.

\begin{table}[!h]
\centering
\small
  \begin{tabular}{cccccc}
    \hline
    \hline
    Dataset & \#Total & \#Training &\#Testing \\
    \hline
    \textbf{KP20k} & 567,830  & 527,830 & 20,000 \\
    \textbf{Krapivin} & 2,304 & 1904 & 400 \\
    \textbf{NUS}  & 211 & - & 211 \\
    \hline
\end{tabular}
{\caption{The statistics of three datasets.}  \label{tab4}}
\vspace{-2mm}
\end{table}

\subsection{Implementation Details}
Due to the limitations of time and GPU resources, all of our models are built on the $\text{BERT}_{base}$ model. All the models are trained on 3 GTX 1080Ti GPUs. During training and testing procedures, the maximum length of the document is 512. 
Moreover, we convert all the documents into lowercase and replace digits with token \emph{$<$digit$>$}. We train our model using an Adam optimizer with a learning rate of 0.001, $ \beta_1= 0.9$, $\beta_2 = 0.998$ and $\epsilon = 10^{-9}$. A dropout rate of 0.1 is applied to avoid overfitting. Gradient clipping is utilized with a maximum norm of 2.0. For PKE, the dimension of the BiLSTM hidden states is set to 512, and we use a linear warm-up strategy with 1000 warm-up steps. For AKG, the encoder and decoder of the Transformer model are all composed of $L=4$ layers, $H=768$ hidden size and $A=8$ attention heads. The warm-up step is set to 8000. For the evaluation of absent keyphrase generation, the beam size is set to 200 on three datasets. We set beam depth to 6. We implement our model with OpenNMT and the Pytorch implementation of BERT. Our code will be released on GitHub.

\subsection{Baseline Models and Evaluation Metrics}

We compare our models with three extractive algorithms (Tf-Idf, TextRank, BiLSTM-CRF~\cite{Alzaidy:2019:BSL:3308558.3313642}) and six state-of-the-art generative baselines, including CopyRNN~\cite{DBLP:conf/acl/MengZHHBC17}, TG-Net~\cite{DBLP:conf/aaai/ChenGZKL19}, KG-KE-KR~\cite{chen-etal-2019-integrated}, CatSeqTG-2RF~\cite{chan-etal-2019-neural}, KG-GAN~\cite{DBLP:journals/corr/abs-1909-12229} and $\text{ParaNet}_T$+CoAtt~\cite{zhao-zhang-2019-incorporating}. 
Following the previous works~\cite{DBLP:conf/aaai/ChenGZKL19,DBLP:conf/acl/MengZHHBC17}, we adopt the macro-averaged \emph{precision}, \emph{recall} and \emph{F-measure} ($F_1$) as evaluation metrics. In the present extraction task, the generation model using beam search ranks the results, while the results of the labeling model are consistent with their original position in the document.
Therefore, the first $k$ keyphrases extracted by labeling models cannot be directly used to calculate $F@k$. We use $F_1@M$ as the evaluation metric following~\cite{chan-etal-2019-neural}, where $F_1@M$ computes an $F_1$ score by comparing all the keyphrase predictions with the ground-truth, i.e., $k$ = the number of predictions.

\subsection{Main Results and Analysis}
In this section, we evaluate our model on two subtasks, i.e., PKE and AKG. We conduct experiments to demonstrate the effectiveness of our approach by comparing it with several state-of-the-art methods on three benchmark datasets. 

\begin{table*}
  \centering
  \begin{tabular}{l|cc|cc|cc}
    \hline
    \hline
    \multirow{2}*{\textbf{Model}} & \multicolumn{2}{|c} {\textbf{KP20k}} & \multicolumn{2}{|c} {\textbf{NUS}} & \multicolumn{2}{|c} {\textbf{Krapivin}}  \\
  & $F_1$@5 & $F_1$@10 & $F_1$@5 & $F_1$@10 & $F_1$@5 & $F_1$@10 \\
  \hline
  \hline
  TF-IDF & 0.105 & 0.130 & 0.139 & 0.181 & 0.113 & 0.143 \\
  TextRank & 0.180 & 0.150 & 0.195 & 0.190 & 0.172 & 0.147 \\
  \hline
  CopyRNN~\cite{DBLP:conf/acl/MengZHHBC17} & 0.378 & 0.310 & 0.418 & 0.369 & 0.339 & 0.281 \\ %
  TG-Net~\cite{DBLP:conf/aaai/ChenGZKL19} & 0.386 & 0.321 & 0.425 & 0.368 &  0.356 & 0.289 \\ %
  \text{KG-KE-KR}~\cite{chen-etal-2019-integrated} & 0.395 & 0.325 & 0.421 & 0.377 & 0.355 & 0.287 \\ 
  CatSeqTG-2RF~\cite{chan-etal-2019-neural} & 0.385 & - & 0.422  & - & 0.369 & - \\ %
  $\text{ParaNet}_T$+CoAtt~\cite{zhao-zhang-2019-incorporating} & 0.360 & 0.289 & 0.360  & 0.350 & 0.329 & 0.282  \\ %
  KG-GAN~\cite{DBLP:journals/corr/abs-1909-12229}  & 0.370  & - & 0.401 & - & 0.357 & - \\ %
  \hline
  BiLSTM-CRF~\cite{Alzaidy:2019:BSL:3308558.3313642}$^\dagger$ & \multicolumn{2}{|c} {0.335} & \multicolumn{2}{|c} {0.351} & \multicolumn{2}{|c} {0.316} \\
   $\text{BERT-PKE}^{\dagger}$ & \multicolumn{2}{|c} {\textbf{0.437}} & \multicolumn{2}{|c} {\textbf{0.447}} &  \multicolumn{2}{|c} {\textbf{0.407}} \\
  \hline
  \end{tabular}
  {\caption{The performance of present keyphrase prediction on three testing datasets. 
  We highlight the best results in bold. $^\dagger$We report the result of $F_1@M$ for the sequence labeling models.} \label{pre}}
\end{table*}

\subsubsection{Present Keyphrase Extraction}

Table~\ref{pre} presents the results of present keyphrase extraction on three datasets. We find that our proposed model outperforms all the generative models and conventional extractive models by a large margin, and our model achieves the highest scores in all test datasets. In particular, our model achieves an improvement of 15.6\%, 13.2\% and 10.6\% over the state-of-the-art model “CopyRNN”, “TG-Net”, and “KG-KE-KR” in the large-scale dataset $\textbf{KP20k}$ respectively. Although the generative models have strong generative capabilities and are assisted with the copying mechanism, the results still show that our model is more capable of extracting present keyphrases from the original document.

\subsubsection{Absent Keyphrase Generation}
The results of the absent keyphrase generation measured by $R@50$ is shown in Table~\ref{abs}.
The results reveal that our model consistently outperforms other baseline methods in all the test datasets again. 
For example, our model achieves an improvement of 24.7\% and 13.2\% over “TG-Net” and “KG-KE-KR,” respectively in $\textbf{KP20k}$.
Besides, the vanilla Transformer model performs better than RNN-based models. 
Note that our fusion module effectively integrates present keyphrase information and BERT knowledge into the Transformer model, bringing substantial improvements on the model performance.

\begin{table}
  \small
  \centering
  \resizebox{1.0\columnwidth}!{
  \begin{tabular}{l|c|c|c}
    \hline
    \hline
    \textbf{Model} & \textbf{KP20k} & \textbf{NUS} & \textbf{Krapivin} \\
   \hline
   \hline
 CopyRNN~\cite{DBLP:conf/acl/MengZHHBC17} & 0.222  & 0.175  & 0.202  \\ %
 TG-Net~\cite{DBLP:conf/aaai/ChenGZKL19}  & 0.226  & 0.164  & 0.169  \\ %
 $\text{KG-KE-KR}$~\cite{chen-etal-2019-integrated}  & 0.249  & 0.190  & 0.252  \\ %
 CatSeqTG-2RF~\cite{chan-etal-2019-neural} & 0.029  & 0.026 & 0.044  \\ %
 $\text{ParaNet}_T$+CoAtt~\cite{zhao-zhang-2019-incorporating} & 0.228  & 0.125   & 0.214  \\ %
 KG-GAN~\cite{DBLP:journals/corr/abs-1909-12229} & 0.027  & 0.027  & 0.037  \\ %
  \hline
  $\text{Transformer}$ & 0.262 & 0.188 & 0.241 \\
  $\text{BERT-AKG}$ & \textbf{0.282}  & \textbf{0.219} & \textbf{0.268} \\
 \hline
  \end{tabular}}
    {\caption{The performance ($R@50$) of absent keyphrase prediction on three testing datasets. 
    We highlight the best results in bold. \label{abs}}}
\end{table}

\begin{table}[t]
\small
\centering
\resizebox{1.0\columnwidth}!{
  \begin{tabular}{l|c|c|c}
    \hline
    \hline
    \textbf{Model} & \textbf{KP20k} & \textbf{NUS} & \textbf{Krapivin} \\
    \hline
    \multicolumn{4}{c} {Extractive}  \\
    \hline
    \text{BERT-PKE} & \textbf{0.437} & \textbf{0.447} & \textbf{0.407} \\
    \quad w/o Sentence Filter & 0.425 & 0.438 & 0.391 \\
    \quad Replace CRF with Linear & 0.350 & 0.339 & 0.267 \\
    \hline
    \multicolumn{4}{c} {Generative}  \\
    \hline
    \text{BERT-AKG} (Fine-tuned$\rightarrow$Fixed BERT) & \textbf{0.282} & \textbf{0.219} & \textbf{0.268} \\
    \quad w/o BERT & 0.262 & 0.188 & 0.241 \\
    \quad w/o Fusion Attention & 0.257 & 0.151 & 0.254 \\
    \quad Fine-tuned$\rightarrow$Fine-tuned BERT & 0.268 & 0.182 & 0.257 \\ 
    \quad Original BERT & 0.278 & 0.209 & 0.264 \\
    \hline
  \end{tabular}
  }
  {\caption{Ablation analysis of our extractive and generative approach on three testing datasets. 
  “Fine-tuned$\rightarrow$Fixed BERT” means that during the training of the generative model, we fix the parameters of BERT which has been fine-tuned on PKE.
  “Fine-tuned$\rightarrow$Fine-tuned BERT” means that we train the BERT fine-tuned on PKE, together with the Transformer model.
  “Original BERT” denotes replacing the BERT fine-tuned on PKE with an original BERT which is not fine-tuned.}
  \label{ablation}}
  \vspace{-3mm}
\end{table}

\subsection{Ablation Study}

Here we conduct some ablation studies for PKE and AKG to explore the effectiveness of our proposed methods. The relevant results of our models are shown in Table~\ref{ablation}.
\paragraph{PKE Ablation.}
Table~\ref{ablation} illustrates that removing our sentence filter module results in a significant decline in the performance. This suggests that our sentence filter module is a core component of our model, which can ameliorate the influences of noisy sentences to improve the labeling ability of the model further.
Besides, we find that the performance degrades by a large margin if the CRF module is replaced with a Linear tagging layer. This indicates that the CRF architecture is crucial for capturing the label dependencies in keyphrase extraction scenarios.

\paragraph{AKG Ablation.} 
As shown in Table~\ref{ablation}, “Fine-tuned$\rightarrow$Fine-tuned BERT” shows that using a fixed BERT as contextual features is better than fine-tuning it on AKG. This may be because excessive training makes BERT forget the useful knowledge obtained from PKE.
The “w/o Fusion attention” result illustrates that our proposed fusion attention module can better integrate the useful information than simply merge them by a weighted average.
Besides, removing the BERT encoder deteriorates the performance, which demonstrates that the present keyphrase information and BERT knowledge can facilitate the generation of absent keyphrases.
Furthermore, we find that replacing our BERT fine-tuned on PKE with an original BERT, which is never fine-tuned, also results in a worse performance, which implies that the extractive task can provide useful information to promote the performance of the generative task.

\begin{figure}[!ht]
    \centering
    \small
    \includegraphics[width=6.5cm, height=4cm]{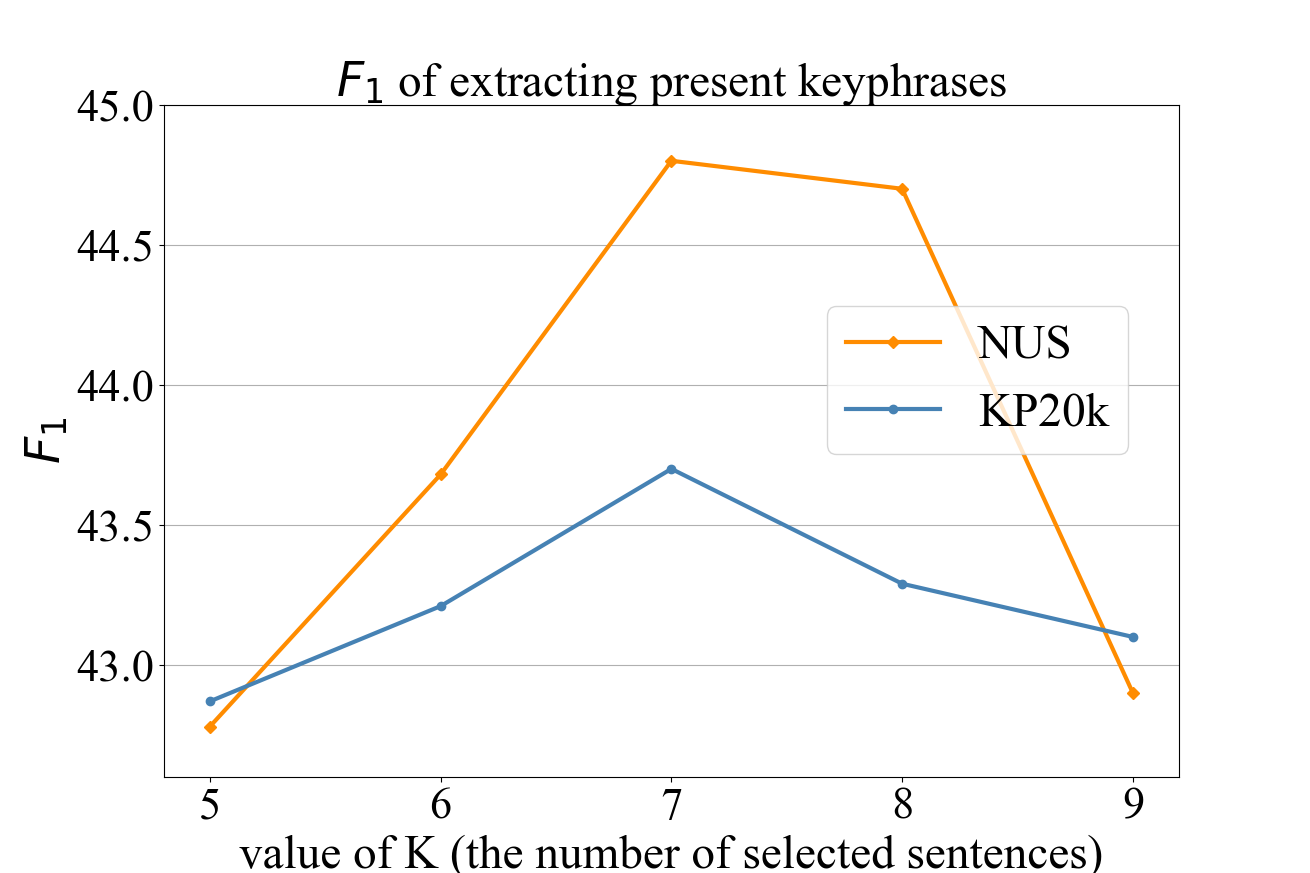}
    \caption{The influence of the number of the selected sentences ($F_1$ of predicting present keyphrases).}
    \label{sel_num}
\end{figure}

\begin{figure*}[ht]\small
\centering
\begin{tabular}{|p{17.1cm}|}
\hline
(1) \ul{\textbf{Macrophages}, \textbf{Oxidation}, and \textbf{Endometriosis}. }
(2) \ul{Retrograde menstruation has been suggested to be the cause for the presence of \textbf{endometrial cells} in the peritoneal cavity.}
(3) However, little is known about the events that lead to the adhesion and growth of these cells that ultimately ... women despite the common occurrence of retrograde menstruation in most women. 
(4) \ul{We postulate that, in normal women, the \textbf{endometrial cells} ... resident tissue macrophages in the peritoneal cavity. }
(5) In contrast, the peritoneal macrophages in women with endometriosis are nonadherent and ineffectively scavenged, resulting in the sustained presence and growth of the endometrial cells. 
(6) \ul{We also postulate that the \textbf{peritoneal fluid} is not a passive reservoir ... , but actively promotes \textbf{endometriosis}. }
(7) \ul{The \textbf{peritoneal fluid} is rich in \textbf{lipoproteins}, particularly low density \textbf{lipoprotein}, which generates oxidized...} 
(8) The oxidants exacerbate the growth of endometriosis by inducing chemoattractants such as mcp ...
(9) We provide evidence for the presence of oxidative milieu in the peritoneal cavity of women with endometriosis, the nonscavenging properties of macrophages that are nonadherent, and the synergistic interaction ...
(10) \ul{For example, the \textbf{peritoneal fluid} \textbf{lipoproteins} of subjects with endometriosis have increased the propensity to undergo \textbf{oxidation} as compared with plasma \textbf{lipoproteins}, ...}
(11) \ul{If the oxidative proinflammatory nature of the \textbf{peritoneal fluid} is an important mediator of \textbf{endometriosis} growth, ... against \textbf{endometriosis}.} \\ 
\specialrule{0.0em}{0pt}{2pt} 
\hline 
\\
\specialrule{0.0em}{0pt}{-8pt}
Present Keyphrases: \quad \{macrophages; oxidation; endometriosis; endometrial cells; peritoneal fluid; lipoproteins\} \\
\enskip \textbf{TG-Net}: \textbf{1. endometriosis}; 2. antioxidants; \textbf{3. macrophages}; \textbf{4. oxidation}; 5. oxidative stress; 6. menstruation; ... ; 10. women \\
\enskip \textbf{KG-KE-KR}: \textbf{1. endometriosis}; 2. antioxidants; \textbf{3. macrophages}; \textbf{4. oxidation}; 5. oxidative stress; ... ; \textbf{10. peritoneal fluid};  \\
\enskip \textbf{BERT-PKE}: \textbf{1. macrophages}; \textbf{2. oxidation}; \textbf{3. endometriosis}; 4. retrograde menstruation; \textbf{5. endometrial cells}; \textbf{6. peritoneal fluid}
\\

\hline
\\
\specialrule{0.0em}{0pt}{-8pt}

Absent Keyphrases: \quad \{growth factors; cytokines\} \\
\enskip \textbf{TG-Net:} \quad~~~1. inflammation; 2. apoptosis; 3. mitochondria; 4. autoimmunity; 5. retrograde oxidation; ... ; \textbf{12. cytokines}; ...\\
\enskip \textbf{KG-KE-KR}: ~1. inflammation; 2. nonadherent scavenged; 3. nonscavenging scavenged; 4. apoptosis; 
...; \textbf{8. cytokines}; ... \\
\enskip \textbf{BERT-AKG}: ~1. inflammation; 2. endometrial growth; 3. retrograde macrophages; \textbf{4. cytokines}; ... ; \textbf{11. growth factors}; ... \\

\hline
\end{tabular}
\caption{Examples of the generated keyphrases by our approach and other models. The underlined sentences in blue are selected by our model. Phrases in bold are true keyphrases and we omit some incorrect predicted keyphrases for brevity. In PKE task, our labeling model only extracts six keyphrases, but for the other two baseline models, we exhibits the top 10 results.}

\label{case}
\end{figure*}

\subsection{Impact of the Number of Selected Sentences} \label{impact}
To further investigate the influence of the hyperparameter \textit{K} (i.e., the number of the selected sentences), we plot $F_1$ curve with respect to different choices of \textit{K}. 
As shown in Figure~\ref{sel_num}, we notice that the curve first rises and then declines with the increase of \textit{K}.
The possible explanation for this phenomenon might be that the recall and precision of the positive sentences reach a balance when \textit{K}=7. 
In other words, the recall of positive sentences increases as the number of selected sentences increases, but the precision declines accordingly.
Hence, the noisy data in the selected sentences is also increasing, leading to the error of labeling.
There is also a consistent trend in the other two datasets.

\subsection{Case Study}

Figure~\ref{case} presents a case of the input document and outputs of different methods. For convenience, we only choose two strong baselines (i.e., TG-Net and KG-KE-KR) for comparison. 
For PKE, our model can decide the appropriate number of keyphrases to be predicted, while the most conventional generative approaches need to select a fixed number of top-ranked candidates as the final results. According to Figure~\ref{case}, our model only extracts six candidates, five of which are correct answers. In comparison, the two baseline models select top-10 keyphrases among which only three are correct.
Furthermore, all the sentences we selected contain present keyphrases. It indicates that our sentence filter can effectively choose positive sentences as much as possible to alleviate the effect of the noisy sentences.
For AKG, all the ground-truth absent keyphrases are included in the results predicted by our model, while the two RNN-based models and the vanilla Transformer model only predict one of them. We observe that our model assigns the keyphrase “cytokines” a higher rank compared with other models. With the help of the present keyphrase knowledge, our model is capable of generating absent keyphrase more accurately.

\section{Related Work}
\paragraph{Keyphrase Extraction.} Extractive methods aim at extracting present keyphrases from the document. 
In most unsupervised methods~\cite{DBLP:conf/aaai/WanX08,DBLP:conf/emnlp/MihalceaT04,DBLP:conf/emnlp/MedelyanFW09}, they first constructed lots of candidate phrases based on some heuristic methods. Then, these candidate phrases were ranked to select those phrases with high scores as the final results.
\citeauthor{DBLP:conf/emnlp/ZhangWGH16}~\shortcite{DBLP:conf/emnlp/ZhangWGH16} and~\citeauthor{Alzaidy:2019:BSL:3308558.3313642}~\shortcite{Alzaidy:2019:BSL:3308558.3313642} tackled the extractive task as a sequence labeling problem. \citeauthor{DBLP:conf/sigir/SunTDDN19}~\shortcite{DBLP:conf/sigir/SunTDDN19} and~\citeauthor{prasad-kan-2019-glocal}~\shortcite{prasad-kan-2019-glocal} adopted the Graph Neural Networks to extract keyphrases by encoding the graph of a document. However, a common drawback is that only relying on these extractive systems can not generate absent keyphrases.

\paragraph{Keyphrase Generation.} Generative methods make it possible to generate absent keyphrase by modeling the keyphrase prediction task as a sequence-to-sequence learning problem. \citeauthor{DBLP:conf/acl/MengZHHBC17} \shortcite{DBLP:conf/acl/MengZHHBC17} first built CopyRNN, a seq2seq framework with attention and copying mechanism~\cite{see-etal-2017-get}. Subsequently, many variations of CopyRNN appeared. \citeauthor{DBLP:conf/emnlp/YeW18} \shortcite{DBLP:conf/emnlp/YeW18} investigated a semi-supervised method for exploitation of the unlabeled data. CorrRNN~\cite{DBLP:conf/emnlp/ChenZ0YL18} employed a review mechanism to the correlation among keyphrases. TG-Net~\cite{DBLP:conf/aaai/ChenGZKL19} leveraged the information of the title to identify the important parts of the document. \citeauthor{chen-etal-2019-integrated} \shortcite{chen-etal-2019-integrated} focused on using an extractive model to enhance the performance of another generative model. \citeauthor{wang-etal-2019-topic-aware}\shortcite{wang-etal-2019-topic-aware} utilized the latent topics of the document to enrich useful features. \citeauthor{zhao-zhang-2019-incorporating} \shortcite{zhao-zhang-2019-incorporating} exploited linguistic constraints to prevent the model from generating overlapping phrases. However, all of them cannot break through the limitation of their generating ability to present keyphrase prediction.

\section{Conclusion}

In this study, we divide the keyphrase prediction into two subtasks: PKE and AKG. 
We introduce a novel joint inference framework to make the most of the power of extractive and generative models.
Specifically, we apply a shared BERT in the two subtasks to make full use of the prior knowledge from the pre-trained language model and share useful information between two subtasks.
The proposed generative model employs the gated fusion attention module to effectively incorporate the updated BERT and Transformer model for better performance on AKG.
The experimental results demonstrate that our approach outperforms the state-of-the-art methods on both PKE and absent AKG tasks.

\bibliographystyle{named}
\bibliography{keyphrase}

\end{document}